\documentclass[conference]{IEEEtran}
\IEEEoverridecommandlockouts
\usepackage{array}
\usepackage{hyperref}
\usepackage{newtxtext}
\usepackage{cite}
\usepackage{amsmath,amssymb,amsfonts}
\usepackage{algorithmic}
\usepackage{graphicx}
\usepackage{textcomp}
\usepackage{xcolor}
\usepackage{multirow}

\def\BibTeX{{\rm B\kern-.05em{\sc i\kern-.025em b}\kern-.08em
    T\kern-.1667em\lower.7ex\hbox{E}\kern-.125emX}}

\makeatletter
\newcommand{\linebreakand}{%
  \end{@IEEEauthorhalign}
  \hfill\mbox{}\par
  \mbox{}\hfill\begin{@IEEEauthorhalign}
}

\begin{document}
\nocite{*}
\title{A Data Selection Approach for Enhancing Low Resource Machine Translation Using Cross-Lingual Sentence Representations\\

}

\author{
\IEEEauthorblockN{Nidhi Kowtal\textsuperscript{ * }}
\IEEEauthorblockA{
\textit{SCTR's Pune Institute of Computer Technology}\\
Pune, India \\
kowtalnidhi@gmail.com}
\hspace{-0.5cm}
\and
\IEEEauthorblockN{Tejas Deshpande\textsuperscript{ * }}
\IEEEauthorblockA{
\textit{SCTR's Pune Institute of Computer Technology}\\
Pune, India \\
tejasdeshpande1112@gmail.com}
 \linebreakand

\IEEEauthorblockN{Raviraj Joshi}
\IEEEauthorblockA{
\textit{Indian Institute of Technology Madras, India}\\
\textit{L3Cube Labs, Pune} \\
Pune, India \\
ravirajoshi@gmail.com}
}

\maketitle
\IEEEpubid{\begin{minipage}{\textwidth}\ \\[20pt] \hspace*{-\parindent}\vspace*{-30pt}\footnotesize * Authors contributed equally.
\end{minipage}}

\begin{abstract}

Machine translation in low-resource language pairs faces significant challenges due to the scarcity of parallel corpora and linguistic resources. This study focuses on the case of English-Marathi language pairs, where existing datasets are notably noisy, impeding the performance of machine translation models. To mitigate the impact of data quality issues, we propose a data filtering approach based on cross-lingual sentence representations.

Our methodology leverages a multilingual SBERT model to filter out problematic translations in the training data. Specifically, we employ an IndicSBERT similarity model to assess the semantic equivalence between original and translated sentences, allowing us to retain linguistically correct translations while discarding instances with substantial deviations. The results demonstrate a significant improvement in translation quality over the baseline post-filtering with IndicSBERT. This illustrates how cross-lingual sentence representations can reduce errors in machine translation scenarios with limited resources.
By integrating multilingual sentence BERT models into the translation pipeline, this research contributes to advancing machine translation techniques in low-resource environments. The proposed method not only addresses the challenges in English-Marathi language pairs but also provides a valuable framework for enhancing translation quality in other low-resource language translation tasks.

\end{abstract}


\section{Introduction and Motivation}
Machine Translation in low-resource language pairs encounters several challenges, with the most significant being the scarcity of parallel corpora and linguistic resources. To overcome these obstacles, datasets are often automatically generated, simplifying the training of translation models. However, datasets produced through these techniques often contain inherent noise, presenting a significant challenge to the creation of reliable and accurate translations. While translating from a high-level language to a low-level language, often it is found there are grammatical errors, and a few words are skipped while translating. This can contribute to noise in these datasets. \newline

\begin{figure}[htbp]
  \centering
  \includegraphics[width=0.5\textwidth]{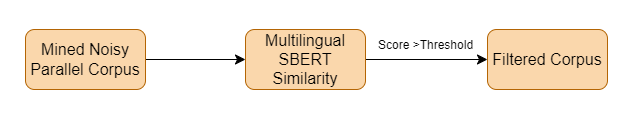}
  \caption{Process of Filtering the Data}
  \label{fig:introduction}
\end{figure}

A few of the methods of automated dataset generation include - 

\textbf{Parallel Corpora Extraction:}
In parallel corpora extraction, texts in multiple languages are aligned using algorithms to match sentences for training translation models. These texts are usually sourced from translated books, articles, or official records. The scarcity of high-quality parallel corpora, particularly for specific language pairs, restricts the effectiveness of this approach and may hinder the size and diversity of the training dataset.

\textbf{Back Transition:}
Utilizing an established translation model, back translation generates a synthetic parallel dataset by translating monolingual data back into the original language. The model's generalization is influenced by potential artefacts and biases introduced from either the translation model or the original training set. Despite this, the method proves effective, especially in scenarios where parallel corpora in the target language pair are limited.

\textbf{Data Augmentation:}
By employing automated methods like paraphrasing, data augmentation can be achieved to enhance the diversity of training data. This aids the model in learning from a broader spectrum of examples, thereby increasing its proficiency in handling linguistic variations. However, excessive and aggressive data augmentation may lead to the generation of nonsensical examples, causing confusion for the model rather than contributing to its generalization.

\textbf{Web Scraping:}
Text content in multiple languages is extracted from online sources using web scraping. Through the use of automated data extraction from websites, we can gather a wide range of texts. To be more precise, we use web scraping to gather parallel translations that are accessible on multilingual websites. Through our website navigation, we can extract sentences that align in different languages, thereby building useful parallel translation corpora. These corpora, which comprise similar sentences in several languages, turn into an essential tool for machine translation model evaluation and training.

\textbf{Pretrained Embeddings:}
The pre-trained embedding method makes use of embeddings produced by pre-existing models that have been trained on linguistic datasets. Semantic relationships between words and sentences are captured by these embeddings. These pre-trained embeddings improve machine translation models' comprehension and representation of linguistic nuances. Through transfer learning, the embeddings help models understand patterns without requiring a lot of task-specific training. By using this technique, the models become more adept at managing a variety of language pairs and translating texts more accurately overall.
\newline

To tackle these issues, we concentrate on the English-Marathi language pair. For machine translation systems, in particular, the noise captures linguistic complexities that present significant challenges. Because of the inherent challenges in generating high-quality translations from these datasets, reliable filtering mechanisms are essential. \newline

Cross-lingual similarity models have been used to construct the datasets under review to replicate the complexities of real-world translation scenarios. Still, we use a stronger model to improve the quality of the filtering process, realizing the need for a more resilient filtering mechanism. The drive to improve translation output accuracy and dependability is driving this switch to a more sophisticated model, which will ultimately help to advance machine translation capabilities in language environments with limited resources. \newline

Our study explores the difficulties that noisy datasets present for low-resource machine translation and highlights the significance of efficient noise reduction techniques. By eliminating problematic sentences, we enhance the quality of translation outputs through the use of the multilingual IndicSBERT model. \newline

\section{Literature Survey}


Machine translation, which has a long history that dates back to the middle of the 20\textsuperscript{th} century, has experienced tremendous evolution over time. The first machine translation attempts were rule-based, translating text between languages using linguistic structures and rules. Unfortunately, these systems had trouble processing the nuances of natural language, which made it difficult to translate words accurately in most cases. \newline

The work \cite{b1} by Vaswani et al. in 2017, is responsible for the development of neural machine translation (NMT) and the popularity of models such as the Transformer architecture. By presenting a self-attention mechanism that could successfully capture complex linguistic patterns and long-range dependencies, the Transformer architecture described in this paper revolutionized the field of machine translation. Using high-resource language pairs, the Transformer architecture's effectiveness was shown, exhibiting notable gains in training efficiency and translation quality. \newline

Machine translation has been impacted by the new era of natural language processing brought about by BERT (Bidirectional Encoder Representations from Transformers). Devlin et al. (2018) \cite{b4} pioneered BERT, a paradigm-shifting approach that pre-trains a deep bidirectional representation of language on an enormous volume of unlabeled text. Researchers have looked into integrating BERT-based models into machine translation architectures. In their investigation into the use of BERT in neural machine translation, Shavarani and Sarkar (2021) concentrated on gathering important linguistic information from BERT to improve the quality of the translated text. Their research highlighted BERT's capacity to identify complex language patterns, enhancing the model's comprehension of semantics and context. \newline

Zhu et al. (2020) \cite{b12} expanded on the investigation of BERT integration into neural machine translation by looking into the advantages of incorporating pre-trained contextual embeddings from BERT. The goal of this study was to improve the representation of source and target language sentences in the translation model by capturing more comprehensive semantic information. The study demonstrated the utility of pre-trained language models in advancing machine translation capabilities by demonstrating the potential of BERT to improve translation accuracy and fluency. \newline

Low-resource language pairs like English to Khasi presented unique challenges that spurred the development of creative methods for achieving efficient machine translation. A Transformer-based method for low-resource neural machine translation from English to Khasi was presented by Thabah and Purkayastha in 2021. \cite{b2} The approach focuses on utilizing the Transformer architecture to improve translation capabilities for underrepresented languages. \newline

In 2021, Gowtham Ramesh \cite{b7} and his associates unveiled the Samanantar project, which aims to solve a persistent problem in machine translation for Indian languages: the lack of parallel corpora. Machine translation models that are trained effectively require parallel corpora, which are collections of aligned texts in multiple languages. Offering the largest collection of parallel corpora for 11 Indian languages that is publicly available, Samanantar stands out as a trailblazing project. To ensure representation from a range of domains, the project sources diverse texts, which are then aligned to create parallel datasets. This comprehensive resource has grown to be essential for scholars and professionals involved in machine translation into the Indian language. Samanantar greatly aids in overcoming the challenges posed by data scarcity by offering a sizable and varied collection of parallel corpora. This makes it possible to develop and assess machine translation models with better language coverage and quality. \newline

\cite{b11} improved the method with semantically weighted back translation for morphologically rich and low-resource languages in the context of unsupervised machine translation. Their study sought to improve unsupervised neural machine translation efficiency while taking into account the unique difficulties presented by the linguistic peculiarities of Indian languages. \newline

One major development in the field was the paradigm shift towards statistical machine translation. Systems such as METEOR and BLEU metrics were developed \cite{b23} for automated assessment, offering numerical values to gauge the calibre of translations. SMT performed well in language pairs with abundant resources, but its low-resource performance—particularly for Indian languages—remained difficult to achieve because of the scarcity of parallel corpora. \newline 

Presented in 2023 by AI4Bharat and partners \cite{b6}, the IndicTrans2 project is an all-encompassing endeavour to fulfil the translation requirements of all 22 scheduled Indian languages. Recognizing India's linguistic diversity, the project seeks to create machine translation models that are both accessible and of excellent quality. To make these models accessible and efficient for all scheduled Indian languages, IndicTrans2 expands on the achievements and difficulties seen in the machine translation field. To emphasize the value of linguistic diversity in the Indian context, the initiative involves the development of specialized models that are suited to the linguistic features of each language. IndicTrans2, with its emphasis on quality and accessibility, stands out as a noteworthy addition to the advancement of machine translation capabilities for all scheduled Indian languages, promoting inclusivity and linguistic representation in the digital sphere. \newline

\section{Methodology}

 \begin{table*}[h]
\centering
\caption{Discrepancies in the Dataset}
  \includegraphics[width=1\textwidth]{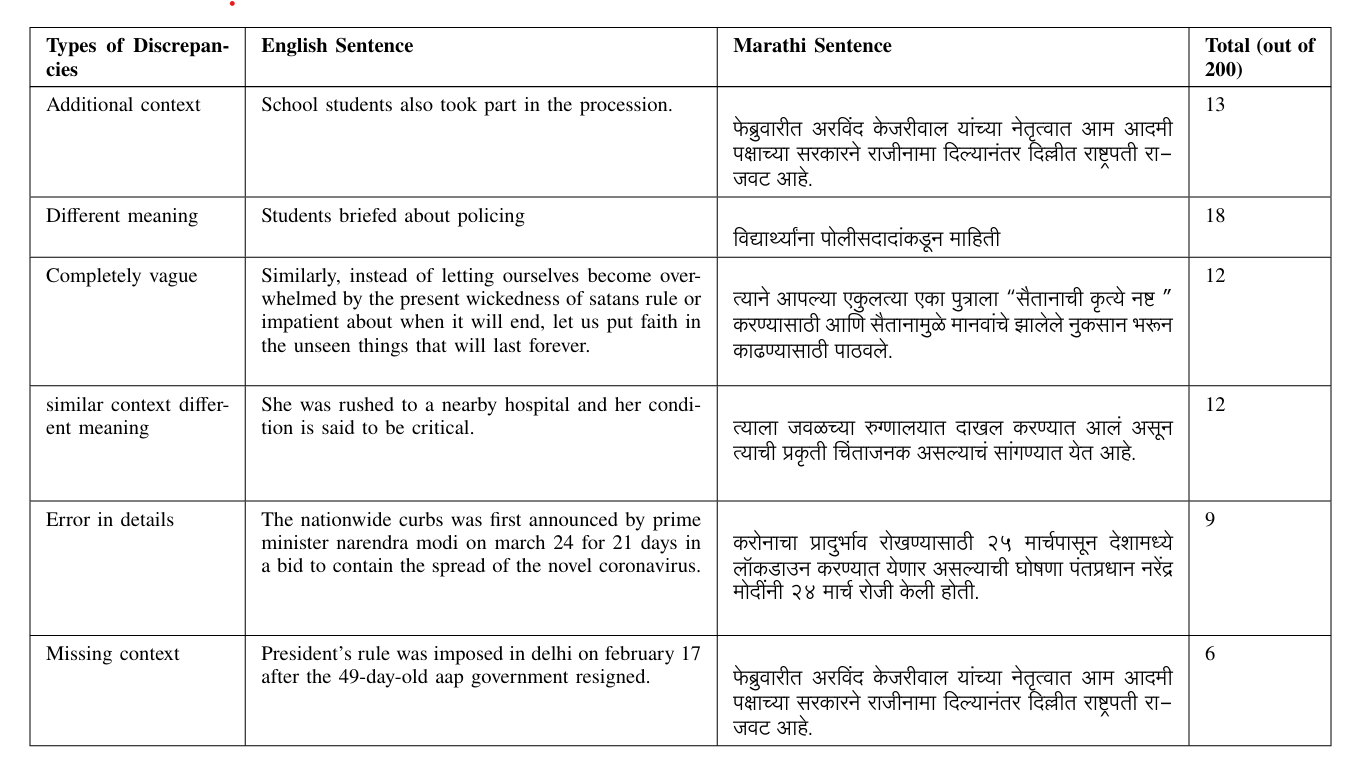}
  \label{tab:discrepancy_dataset}
\end{table*}

\begin{figure*}[htbp]
  \centering
  \includegraphics[width=0.8\textwidth]{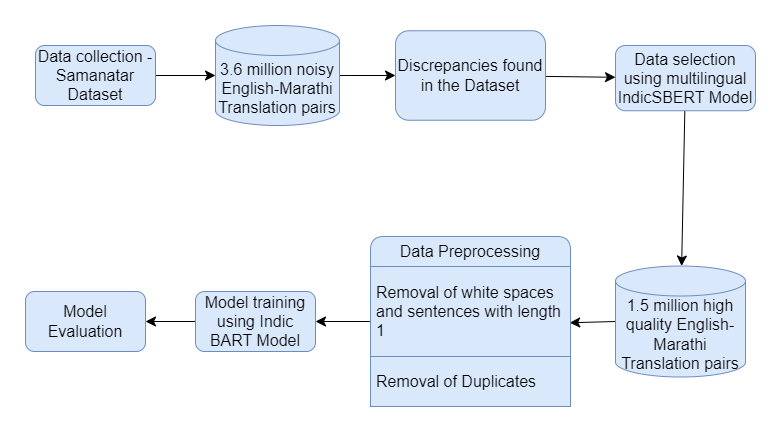}
  \caption{Stages of the proposed approach}
  \label{fig:ict_paper_fig}
\end{figure*}

\subsection{~\ref{fig:ict_paper_fig}Overview of our proposed approach}


\subsubsection{\textbf{Original Data}}

The initial dataset consists of 3.6 million sentences which are the ai4bharat's BPCC Mined Dataset, representing the original noisy data. These sentences are noisy since they are mined and are not manually annotated.
This dataset includes diverse sentences with potential variations in grammar, context, and translation quality.

\subsubsection{\textbf{Discrepancies found in the Dataset}}
After evaluating the 3.6 million corpus, we found out that the dataset contained duplicates. So the sentence pairs which had the same translations were removed, and the language pairs which had different translations in any one language were retained, since the model would get attenuated to different contexts in a language.

In Table~\ref{tab:discrepancy_dataset}, we present the discrepancies found in the dataset.
Following a manual assessment of 200 randomly chosen sample sentences from the dataset, various kinds of differences between the Marathi and English translations were found. These include situations where the Marathi translation did not fully convey subtleties from the English sentence. There have been instances where the translated text conveyed a different meaning due to cases of different meanings. Certain translations lacked specificity and were utterly ambiguous. Inconsistencies also appeared in details and missing contextual information, as well as sentences with similar contexts but distinct meanings.
\newline
The fact that nearly 50\% of the sampled data showed these kinds of inconsistencies is notable and underscores the difficulties in preserving accurate translations across the dataset. 

\subsubsection{\textbf{Data selection using multilingual IndicSBERT}}
The IndicSBERT model ~\ref{fig:introduction} is incorporated into the methodology as a sophisticated tool for measuring sentence similarity. The IndicSBERT model is used to get the similarity score between the English sentence and its corresponding Marathi sentence. The similarity score ranges between 0 to 1. We expelled the sentences whose similarity score was below 0.7. We trained the model only on the filtered sentences with a similarity score greater than 0.7. Thus, we procured high-quality dataset of 1.5 million corpus.\newline

\subsubsection{\textbf{Model Training}}
   We trained our on top of IndicBART model, using 1 Million sentences after pre-processing the dataset. We found some shortcomings in the translations.\\
During the translation process, we noticed that some translations were grammatically incorrect and included English words. After reviewing the dataset, we found it to be noisy with several discrepancies. Since manually correcting such a large dataset wasn't feasible, we decided to use the IndicSBERT model to address these issues.

The IndicSBERT model is incorporated into the methodology as a sophisticated tool for measuring sentence similarity. The IndicSBERT model was used to get the similarity score between the English sentence and its corresponding Marathi sentence. The similarity score ranges between 0 to 1. We expelled the sentences whose similarity score was below 0.7. We trained the model only on the filtered sentences with a similarity score greater than 0.7. The outputs of all three models—the original IndicBART, the fine-tuned IndicBART, and the filtered dataset using IndicSBERT are thus analyzed.

By filtering sentences based on the similarity score, we included 1 million sentences in the training dataset. After training the model on this dataset, the results showed significant improvement compared to the previous model. The translations were grammatically correct and no longer contained any English words.

\subsubsection{\textbf{Model Evaluation}}
  The baseline model and our model are evaluated on the BLEU score, METEOR Score, CHRF Score, CHRF++ Score and IndicSBERT Score, and are mentioned in the table.
  
\subsection{\textbf{Dataset details}}
We chose "AI4Bharat's BPCC" Dataset for our model trainng.
BPCC is an extensive collection of parallel corpora created for eleven different Indic languages. The collection includes parallel texts written in Assamese, Bengali, Marathi, Gujarati, Oriya, Tamil, Telugu, Malayalam, Bengali, Bengal, and Hindi. It is a valuable tool for tackling the problem of sparse parallel corpora for low-resource languages.

\subsection{\textbf{Models:}}
\subsubsection{\textbf{IndicBART Model}}
Specifically designed for Indian languages, the IndicBART model is a transformer-based architecture. It serves a wide linguistic spectrum, having been pre-trained on a multilingual corpus that includes 11 major Indian languages. Perfected for tasks like machine translation from English to Marathi, the model is excellent at interpreting the subtle differences between these two languages. Its proficiency in producing accurate and contextually relevant translations for this particular language is ensured by its specialization in Marathi. It is a useful tool for researchers and developers working on Indian language natural language processing tasks.

 \subsubsection{\textbf{IndicSBERT Model}}
 Indic-Sentence-BERT (SBERT) model optimized for Indian language sentence similarity is created by L3Cube and has been trained on a wide range of Indian languages, such as Bengali, Tamil, Telugu, Kannada, Malayalam, Hindi, Marathi, and Gujarati. It is especially good at capturing semantic relationships between sentences by utilizing the SBERT architecture. This makes it useful for tasks like filtering out dissimilar sentences in a multilingual context. The model is skilled at recognizing the contextual subtleties that affect sentence similarity because it has been fine-tuned for the complexities of Indian languages. Sentences in Hindi, Marathi, or any other supported Indian language can be used to measure the semantic similarity between sentence pairs using the embeddings provided by this model. It provides a practical and approachable way for developers and researchers to include sentence similarity measurement in applications and research projects about Indian languages.

 \begin{table*}[h]
\centering
\caption{Some translations performed by previous model v/s the fine-tuned model}
  \includegraphics[width=1\textwidth]{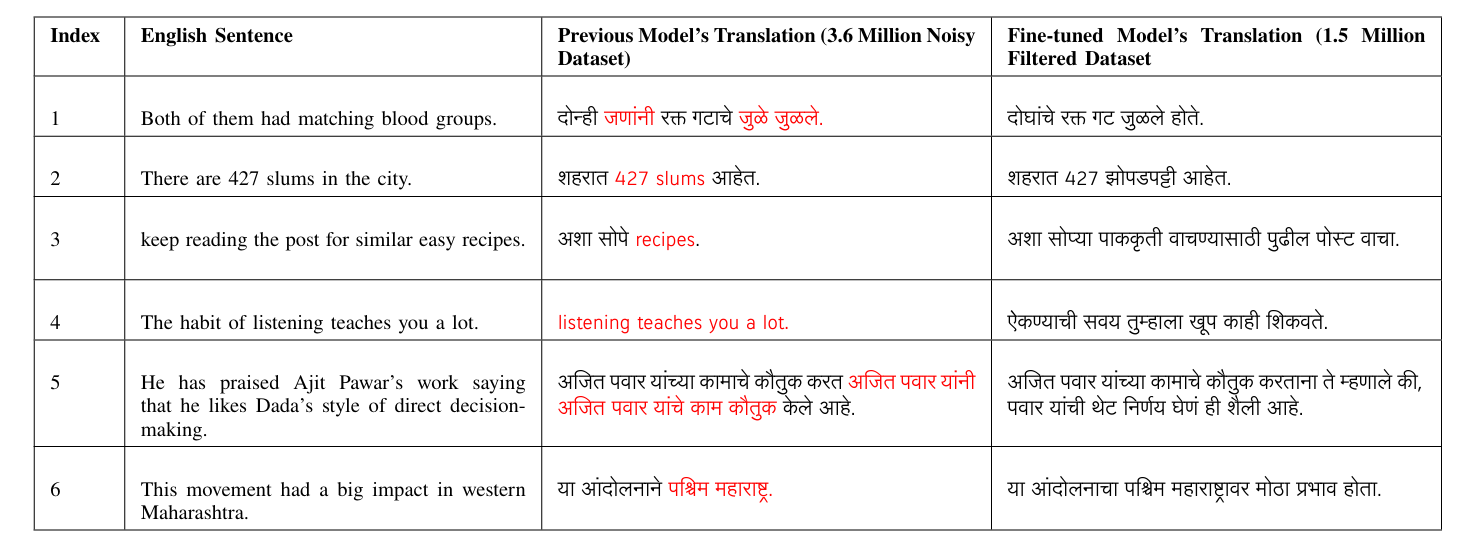}
  \label{tab:results}
\end{table*}

\begin{table*}[h]
\centering
\caption{Evaluation Metrics}
\label{tab:indic_results}
\renewcommand{\arraystretch}{1.5}
\begin{tabular}{|>{\centering\arraybackslash}m{3cm}|
                >{\centering\arraybackslash}m{3cm}|
                >{\centering\arraybackslash}m{3cm}|}

\hline
\textbf{} & \textbf{English to Marathi} & \textbf{English to Marathi} \\
\hline
\textbf{Metrics} & \textbf{Mean value of metric before filtering the dataset} & \textbf{Mean value of metric after filtering the dataset} \\
\hline
\multirow{2}{*}{IndicSBERT Score} & 75.3 & 78.2   \\
& &  \\
\hline
\multirow{2}{*}{BLEU Score} & 28.9 & 35.6  \\
& &  \\
\hline
\multirow{2}{*}{Meteor Score} & 28.7 & 34.1  \\
& &  \\
\hline
\multirow{2}{*}{CHRF Score} & 37.044 & 43.674  \\
& &  \\
\hline
\multirow{2}{*}{CHRF++ Score} & 37.860 & 44.223  \\
& &  \\
\hline

\end{tabular}
\end{table*}

\section{Gold Testset Curation}
It was observed that the dataset which was being used as a test dataset previously had some errors in the translation. So we considered the news dataset available on the MahaNLP corpus. Initially, we randomly considered 10000 sentences from the dataset. Then we manually went through these 10000 sentences and selected the top 1500 sentences which were translated accurately. We then calculated the metrics of our translation model using this manually curated test dataset. To check our model's performance, we have considered the following metrics.

\section{Results and Discussion}
\subsection{Evaluation Metrics} 
\begin{enumerate}
    \item IndicSBERT Score - 
    \newline 
    The IndicSBERT model helps to check the similarity score between translated sentences with their respective Marathi sentence from the dataset. Our fine-tuned model are evaluated on a manually curated test dataset, using the IndicSBERT score as a metric. The mean IndicSBERT score of both the models is given in ~\ref{tab:indic_results} table.
    \newline

    \item BLEU Score -
    \newline
    The BLEU Score is the benchmark for evaluating the quality of computer translation. It compares the translations produced by the ML model and the actual translations. It examines word clusters, such as one word, two words together, and so forth. It counts the number of words that are translated the same in both the model and the actual translations for each group. It then determines a score. A score of one indicates flawless translation. If it is zero, then no word was correctly predicted by the model. Higher BLEU Scores indicate that the translation produced by the model is more accurate than the original. Table ~\ref{tab:indic_results} indicates the BLEU score of both models.
    \newline
    
    \item Meteor Score - 
    \newline
    The METEOR score is a metric used to measure the quality of machine-generated translations by comparing them to reference translations, which are considered the gold standard. It considers various aspects like precision, recall, and alignment to evaluate how well the generated translation captures the meaning and nuances of the original text. METEOR is particularly useful in machine translation evaluation because it goes beyond simple word matching and considers the overall fluency and correctness of the translated sentences. Higher METEOR scores indicate more accurate and contextually relevant translations, providing a quantitative measure for assessing the performance of machine translation models. Table ~\ref{tab:indic_results} indicates the Meteor score of both the models.
    \newline
    
    \item CHRF and CHRF++ Score -
    \newline
    These metrics function at the character level, as opposed to conventional metrics that concentrate on words. Essentially, they use matching character n-gram analysis to systematically check if translations produced by machines and humans are consistent.

A higher CHRF score in this case indicates improved performance by showing a good alignment between the machine translation and the human reference. An improved version of CHRF, called CHRF++, expands its analysis to include different character n-gram lengths, offering a more complex evaluation of the translation quality. \newline 

\end{enumerate}

\subsection{Observations from Evaluation Metrics} 

The five metrics mentioned above namely IndicSBERT score, BLEU Score, Meteor Score, CHRF Score and CHRF++ Score were used for the evaluation. The scores ~\ref{tab:indic_results} of all the metrics have improved significantly after training the model on a filtered dataset. IndicSBERT Score increased by 2.\%. BLEU Score improved by 6.7\%. Meteor Score improved by 5.4\%. CHRF Score was increased by 6.63\% and CHRF++ Score was increased from 6.4\%.

\subsection{Observations from Sample Translations} 
Observations from the examples in Table  ~\ref{tab:results}  :\\
1) In the 1st example, the translation of the fine-tuned model is grammatically correct.\\
2) In sentences 2 and 3, the translations of the previous model included some English words, but the fine-tuned model's translation doesn't contain any English words. \\
3) In sentence number 4, the translation of the previous model contains a part of the English sentence as it is, whereas the fine-tuned model's translation is accurate\\
4) In the 5th sentence, the translation of the previous model contains the same name thrice and is not accurate, whereas the translation of the finetuned model is accurate\\
5) In the 6th sentence, the translation of the previous model is incomplete, whereas the translation of the fine-tuned model is complete.



\section{Conclusion and Future Scope}
In this paper, we have highlighted a method to filter the noisy, low-resource dataset. It was concluded that training the model after filtering out the noisy sentences from the dataset improved the performance of the model.
\newline
Our long-term goal is to obtain additional high-quality data for languages with scarce resources, which has been a recurring problem in our work. We want to work with linguists and institutions to gather large, varied datasets so that our models can be trained more effectively.

\section{Acknowledgement}
This work was done under the mentorship of Mr. Raviraj Joshi (Mentor, L3Cube Pune). We would like to express our gratitude towards him for his continuous support and encouragement.

\bibliographystyle{plain}
\bibliography{main}
\end{document}